\documentclass{article} 
\usepackage{iclr2018_conference,times}
\usepackage[colorlinks=true,linkcolor=blue,allcolors=blue]{hyperref}
\usepackage{url}

\usepackage{epsfig}
\usepackage{graphicx}
\usepackage{amsmath}
\usepackage{amssymb}
\usepackage{color}
\usepackage{array, floatrow, tabularx, makecell, booktabs}
\setcellgapes{3pt}
\usepackage{wrapfig}
\usepackage{verbatim}
\usepackage{wrapfig}
\usepackage{graphicx}
\usepackage{caption}
\usepackage{subcaption}

\title{Apprentice: Using Knowledge Distillation Techniques To Improve
Low-Precision Network Accuracy}

\author{Asit Mishra \& Debbie Marr \\
Accelerator Architecture Lab\\
Intel Labs\\
\texttt{\{asit.k.mishra,debbie.marr\}@intel.com} \\
}

\iclrfinalcopy 

\begin{document}

\maketitle

\begin{abstract}

Deep learning networks have achieved state-of-the-art
accuracies on computer vision workloads like image classification
and object detection. The performant systems, however,
typically involve big models with numerous parameters.
Once trained, a challenging aspect for such top performing
models is deployment on resource constrained
inference systems --- the models (often deep networks
or wide networks or both) are compute and memory intensive.
\textit{Low-precision numerics} and model compression
using \textit{knowledge distillation} are popular techniques
to lower both the compute requirements and memory footprint
of these deployed models.
In this paper, we study the \textit{combination} of
these two techniques and show that the performance of
low-precision networks can be \textit{significantly}
improved by using knowledge distillation techniques.
Our approach, \textit{Apprentice}, achieves
state-of-the-art accuracies using ternary precision
and 4-bit precision for variants of ResNet
architecture on ImageNet dataset.
We present three schemes using which one can apply knowledge
distillation techniques to various stages of the
train-and-deploy pipeline.

\end{abstract}

\section{Introduction}

\textbf{Background}: Today's high performing deep neural networks (DNNs) for computer vision
applications comprise of multiple layers and involve numerous parameters.
These networks have $\mathcal{O}($Giga-FLOPS$)$ compute requirements and
generate models which are $\mathcal{O}($Mega-Bytes$)$ in
storage~\citep{Eugenio-DNN-complexity}.
Further, the memory and compute requirements
during training and inference are quite different~\citep{WRPN}. Training
is performed on big datasets with large batch-sizes
where the memory footprint
of activations dominates the model memory footprint.
On the other hand, the batch-size during
inference is typically small
and the model's memory footprint dominates the
runtime memory requirements.

Because of the complexity in compute, memory and storage requirements,
training phase of the networks
is performed on CPU and/or GPU clusters
in a distributed computing environment.
Once trained, a challenging aspect is deployment of the trained models
on resource constrained inference systems such as
portable devices or sensor networks,
and for applications in which real-time predictions are required.
Performing inference on edge-devices comes with severe
constraints on memory, compute and power.
Additionally, ensemble based methods, which one can
potentially use to get
improved accuracy predictions, become prohibitive in
resource constrained systems.

Quantization using low-precision numerics~\citep{GoogleLowPrecision, DoReFa, NN-FewMult,
LogNN, StochasticRounding, TTQ,
XNORNET, BWN, FINN, WRPN} and
model compression~\citep{Compression-Caruana,Hinton-distill, Fitnets}
have emerged as popular solutions for
resource constrained deployment scenarios.
With quantization, a low-precision version of the network model is generated
and deployed on the device. Operating in lower precision mode reduces compute as
well as data movement and storage requirements.
However, the majority of existing works in low-precision
DNNs sacrifice accuracy over the baseline full-precision networks.
With model compression, a smaller low memory footprint network
is trained to mimic the behaviour of the
original complex network. During this training, a process
called, knowledge distillation is used to ``transfer knowledge''
from the complex network to the smaller network.
Work by~\citet{Hinton-distill} shows that the knowledge
distillation scheme can yield
networks at comparable or slightly better accuracy than the original complex
model. However, to the best of our knowledge, all prior
works using model compression techniques target
compression at full-precision.

\textbf{Our proposal}: In this paper, we study the combination of network quantization
with model compression and show that the accuracies of
low-precision networks can be significantly
improved by using knowledge distillation techniques.
Previous studies on
model compression use a large network as the teacher network
and a small network as the student network. The small student network learns from the
teacher network using the distillation process.
The network architecture of the student network
is typically different from that of the teacher network --
for e.g.~\citet{Hinton-distill} investigate a student
network that has fewer number of neurons in the hidden layers compared to the
teacher network. In our work, the student network has similar
topology as that of the teacher network, except that the
student network has low-precision
neurons compared to the teacher network which has neurons
operating at full-precision.

We call our approach \textit{Apprentice}\footnote{Dictionary defines
apprentice as a person
who is learning a trade from a skilled employer,
having agreed to work for a fixed period at low wages. In our work,
the apprentice is a low-precision
network which is learning the knowledge of a high precision
network (skilled employer) during a fixed number of epochs.} and study three schemes
which produce low-precision networks
using knowledge
distillation techniques. Each of these three schemes produce state-of-the-art
ternary precision and 4-bit precision models.

In the first scheme, a low-precision network and a full-precision network are
jointly trained from scratch using knowledge distillation scheme.
Later in the paper we describe the rationale behind this approach.
Using this scheme,
a new state-of-the-art accuracy is obtained for ternary and 4-bit precision
for ResNet-18, ResNet-34 and ResNet-50 on ImageNet dataset.
In fact, using this scheme the accuracy
of the full-precision model also slightly improves.
This scheme then serves as the new baseline for the
other two schemes we investigate.

In the second scheme, we start with a full-precision trained network
and transfer knowledge from this trained network continuously to train a
low-precision network from scratch. We find that the low-precision network converges
faster (albeit to similar accuracies as the first scheme) when a trained
complex network guides its training.

In the third scheme, we start with a trained full-precision large network
and an apprentice network that has been initialised with full-precision weights.
The apprentice network's precision is lowered and is fine-tuned
using knowledge distillation techniques.
We find that the low-precision network's accuracy marginally
improves and surpasses the accuracy obtained via the first scheme. This scheme then
sets the new state-of-the-art accuracies for the ResNet models at ternary and 4-bit
precision.

Overall, the contributions of this paper are the techniques to obtain low-precision DNNs using
knowledge distillation technique. Each of our scheme produces a low-precision model
that surpasses the accuracy of the equivalent low-precision model published to date.
One of our schemes also helps a low-precision model converge faster. We envision these
accurate low-precision models to simplify the inference deployment process on resource
constrained systems and even otherwise on cloud-based deployment systems.

\section{Motivation for low-precision model parameters}

\begin{wrapfigure}{r}{0.45\textwidth}
\begin{center}
   \includegraphics[width=1.0\linewidth]{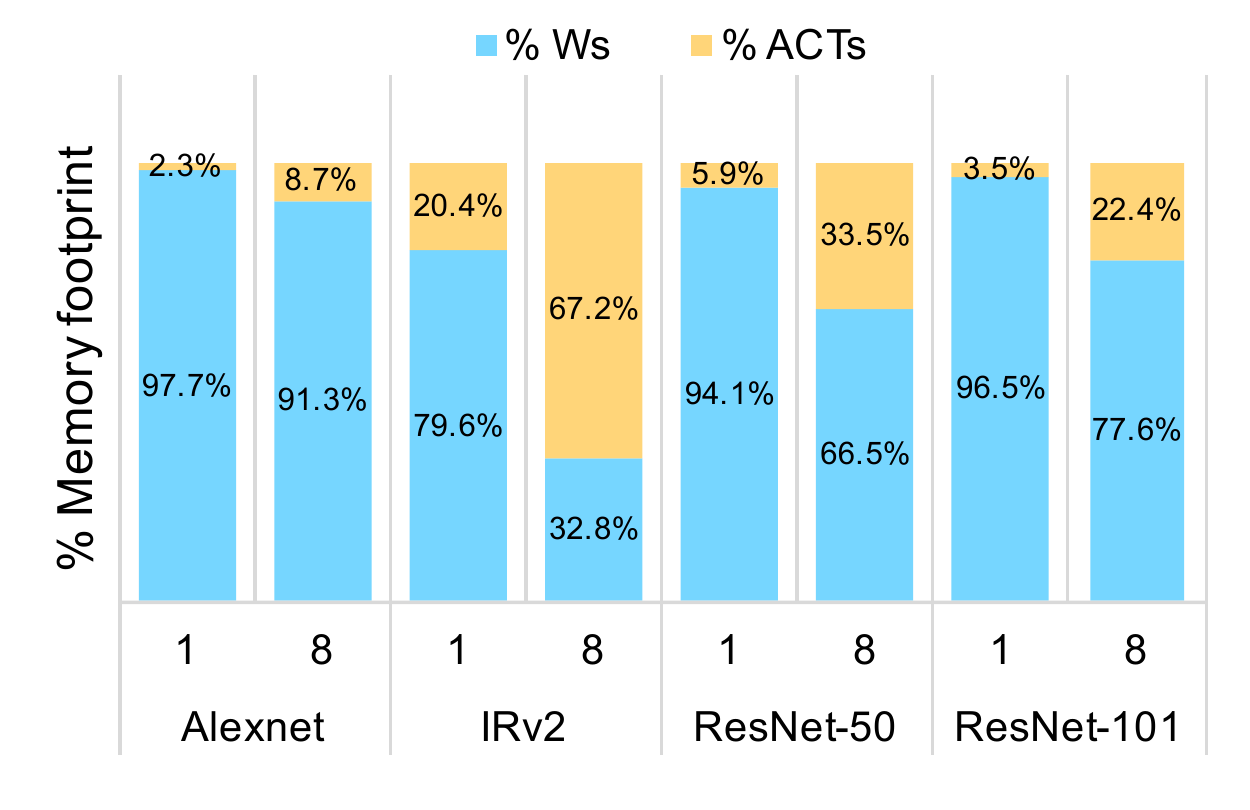}
\end{center}
   \caption{\small Memory footprint of activations (ACTs) and
   weights (W) during inference for mini-batch sizes 1 and 8.}
\label{fig:MemoryFootprint}
\end{wrapfigure}

\textbf{Lowering precision of model parameters}: Resource constrained inference systems impose
significant restrictions on memory, compute and power budget. With regard to storage,
model (or weight) parameters and activation maps occupy memory during the inference phase of DNNs.
During this phase memory is allocated for input (IFM) and output
feature maps (OFM) required by a single layer in the DNN, and these dynamic memory allocations
are reused for other layers. The total memory allocation during inference is then
the maximum of IFM and maximum of OFM memory required across all the layers
plus the sum of all weight tensors~\citep{WRPN}.
When inference phase for DNNs is performed with a small batch size, the memory footprint
of the weights exceeds the footprint of the activation maps. This aspect is shown in
Figure~\ref{fig:MemoryFootprint} for 4 different networks (AlexNet~\citep{AlexNet},
Inception-Resnet-v22~\citep{IRv2}, ResNet-50
and ResNet-101~\citep{ResNet}) running 224x224 image patches. Thus lowering the
precision of the weight tensors
helps lower the memory requirements during deployment.

\textbf{Benefit of low-precision compute}: Low-precision compute simplifies
hardware implementation. For example, the compute unit to
perform the convolution operation (multiplication of two operands) involves a floating-point
multiplier when using full-precision weights and activations.
The floating-point
multiplier can be replaced with a much simpler circuitry (xnor and
popcount logic elements) when using binary precision
for weights and activations~\citep{BNN, XNORNET, BWN}.
Similarly, when using ternary precision
for weights and full-precision
for activations, the multiplier unit can be replaced with a sign comparator
unit~\citep{TWN, TTQ}. Simpler hardware also helps lower
the inference latency and energy
budget. Thus, operating in lower precision mode reduces compute as
well as data movement and storage requirements.

The drawback of low-precision models, however, is degraded accuracy.
We discuss later in the paper
the network accuracies obtained
using methods proposed in literature.
These accuracies serve as the starting point and baselines we compare to in our work.

\section{Related work}

\textbf{Low-precision networks}: Low-precision
DNNs are an active area of research.
Most low-precision networks acknowledge the over parameterization aspect
of today's DNN architectures and/or the aspect that lowering
the precision of neurons post-training often
does not impact the final performance.
Reducing precision of weights for efficient
inference pipeline has been very well studied.
Works like Binary connect (BC)~\citep{BWN}, Ternary-weight
networks (TWN)~\citep{TWN}, fine-grained ternary
quantization~\citep{PCLpaper1} and INQ~\citep{INQ} target
precision reduction of network weights. Accuracy is
almost always affected when quantizing the weights
significantly below 8-bits of precision.
For AlexNet on ImageNet, TWN loses 5\% Top-1 accuracy.
Schemes like INQ, work in~\citet{ResilientDNNs} and \citet{PCLpaper1}
do fine-tuning to quantize the network weights.

Work in XNOR-NET~\citep{XNORNET}, binary neural networks~\citep{BNN}, DoReFa~\citep{DoReFa}
and trained ternary quantization (TTQ)~\citep{TTQ} target training pipeline. While TTQ
targets weight quantization, most works targeting
activation quantization show that quantizing activations always
hurt accuracy. XNOR-NET approach degrades Top-1 accuracy by 12\%
and DoReFa by 8\% when quantizing both weights and activations
to 1-bit (for AlexNet on ImageNet).
Work by~\citet{StochasticRounding} advocates for low-precision fixed-point
numbers for training. They show 16-bits to be sufficient for training
on CIFAR10 dataset.
Work by~\citet{1bGradient} quantizes gradients
in a distributed computing system.

\textbf{Knowledge distillation methods}:
The general technique in distillation based methods involves using a teacher-student strategy,
where a large deep network trained for a given task teaches
shallower student network(s) on the same task.
The core concepts behind knowledge distillation or transfer technique have been around for a while.
\citet{Compression-Caruana} show that one can compress the information in an
ensemble into a single network.~\citet{BaC13} extend this approach
to study shallow, but wide, fully connected topologies by mimicking deep neural networks.
To facilitate learning, the authors introduce the concepts of learning on
logits rather than the probability distribution.

\citet{Hinton-distill} propose a
framework to transfer knowledge by introducing the concept of temperature. The key idea is to
divide the logits by a temperature factor before performing a Softmax function. By
using a higher temperature factor the activations of incorrect classes
are boosted. This then facilitates more information flowing to the model parameters during
back-propagation operation. FitNets~\citep{Fitnets} extend this work by using intermediate hidden layer
outputs as target values for training a deeper, but thinner, student model.
Net2Net~\citep{Net2Net} also uses a teacher-student network system
with a function-preserving transformation approach to initialize the parameters of the
student network. The goal in Net2Net approach is to accelerate the training of a
larger student network.
\citet{AttentionCNN} use attention as a mechanism for transferring knowledge
from one network to another. In a similar theme,~\citet{Yim_2017_CVPR}
propose an information metric
using which a teacher DNN can transfer the distilled knowledge to other student DNNs.
In N2N learning work, ~\citet{N2N} propose a reinforcement learning based approach
for compressing a teacher network into an equally capable student network. They achieve a compression
factor of 10x for ResNet-34 on Cifar datasets.

\textbf{Sparsity and hashing}: Few other popular techniques for model compression are
pruning~\citep{NIPS1989_250, Han-pruning, structured-sparsity, HanPTD},
hashing~\citep{Hash1} and weight sharing~\citep{Hash2, Hash3}.
Pruning leads to removing neurons entirely from the final
trained model making the model a sparse structure. With hashing and weight sharing schemes
a hash function is used to alias several weight parameters into few hash buckets, effectively
lowering the parameter memory footprint.
To realize benefits of sparsity and hashing schemes during runtime, efficient
hardware support is required (e.g. support for
irregular memory accesses~\citep{HanEIE,AALTWN,SCNN}). 

\section{Knowledge Distillation} \label{sec:kd}

We introduce the concept of knowledge distillation in this section.~\citet{Compression-Caruana},
~\citet{Hinton-distill} and~\citet{2016arXiv160305691U} analyze this topic in great detail.

\begin{figure}[!htb]
\begin{center}
   \includegraphics[width=0.75\linewidth]{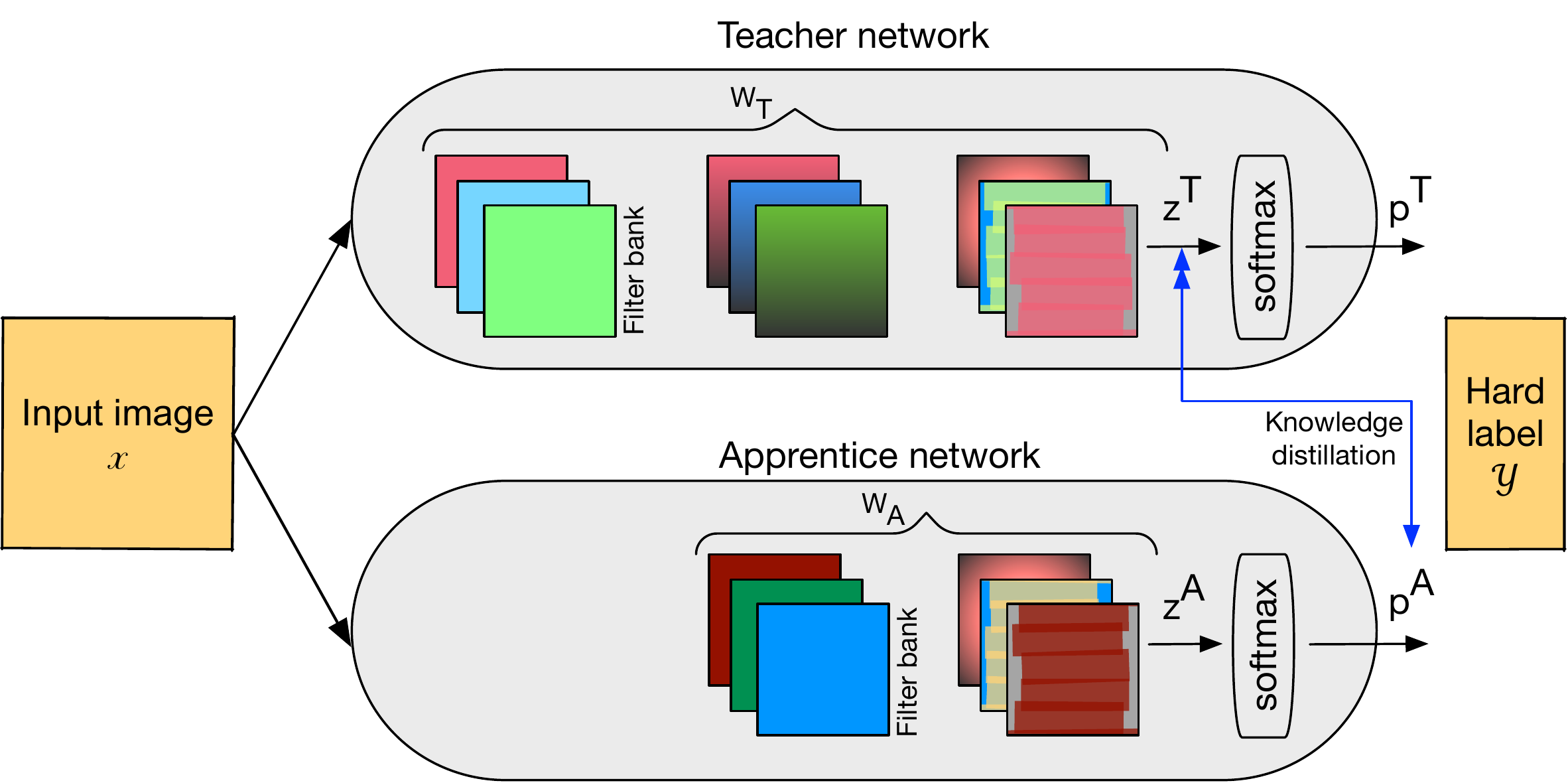}
\end{center}
   \caption{\small Schematic of the knowledge distillation setup. The teacher
   network is a high precision network and the apprentice network is a low-precision network.}
\label{fig:ArchitectureApprenticeFig}
\end{figure}

Figure~\ref{fig:ArchitectureApprenticeFig} shows the schematic of the
knowledge distillation setup.
Given an input image $x$, a teacher DNN maps this image to predictions $p^T$.
The $C$ class predictions are obtained by applying
Softmax function on the un-normalized log probablity values $z$ (the logits), i.e.
$p^T$ = ${ e^{z^{T}_{k}} }/{ \sum_{j}^{C} e^{z^{T}_{j}}}$. The same image is fed to
the student network and it
predicts $p^A$ = ${ e^{z^{A}_{k}} }/{ \sum_{j}^{C} e^{z^{A}_{j}}}$.
During training, the cost function, $\mathcal{L}$,
is given as:

\begin{equation} \label{eq1}
\mathcal{L}(x;W_T,W_A) = \alpha\mathcal{H}(y,p^T) + \beta\mathcal{H}(y,p^A) + \gamma\mathcal{H}(z^T,p^A)
\end{equation}

where, $W_T$ and $W_A$ are the parameters of the teacher
and the student (apprentice) network, respectively, $y$ is the ground truth,
$\mathcal{H}(\cdot)$ denotes a loss function and,
$\alpha$, $\beta$ and $\gamma$ are weighting factors to prioritize the
output of a certain loss function over the other.

In equation~\ref{eq1}, lowering the first term of the cost function
gives a better teacher network and lowering the second
term gives a better student network. The third term is the
knowledge distillation term whereby the student network attempts to
mimic the knowledge in the teacher network. In~\citet{Hinton-distill}, the logits of
the teacher network are divided by a temperature factor $\tau$. Using a higher value
for $\tau$ produces a softer probability distribution when taking the Softmax of the
logits. In our studies, we use cross-entropy function for $\mathcal{H}(\cdot)$, 
set $\alpha=1$, $\beta=0.5$ and $\gamma=0.5$ and, perform the transfer learning process
using the logits (inputs to the Softmax function) of the teacher network.
In our experiments we study the effect of varying the depth of the teacher
and the student network, and the precision of the neurons
in the student network.

\section{Our approach - Apprentice network}

Low-precision DNNs target the storage and compute efficiency aspects of the network.
Model compression targets the same efficiency parameters from the point of view of
network architecture. With \texttt{Apprentice} we combine both these techniques to
improve the network accuracy as well as the runtime efficiency of DNNs.
Using the teacher-student setup described in the last section, we investigate three
schemes using which one can obtain a low-precision model for the
student network.
The first scheme (scheme-A) jointly trains both the networks -
full-precision teacher and low-precision student
network. The second scheme (scheme-B) trains only the low-precision student network
but distills knowledge from a trained full-precision teacher network
throughout the training process.
The third scheme (scheme-C) starts with a trained full-precision teacher
and a full-precision student network
but fine-tunes the student network after lowering its precision.
Before we get into the details of each of these schemes, we discuss the accuracy numbers obtained
using low-precision schemes described in literature.
These accuracy figures serve as the baseline for comparative analysis.

\subsection{Top-1 error with prior proposals for low-precision networks}
We focus on sub 8-bits precision for inference deployments,
specifically ternary and 4-bits
precision.
We found TTQ~\citep{TTQ} scheme achieving the state-of-the-art accuracy
with ternary precision for weights and full-precision (32-bits floating-point)
for activations.
On Imagenet-1K~\citep{ILSVRC15},
TTQ achieves 33.4\% Top-1 error rate with a ResNet-18 model.
We implemented TTQ scheme for ResNet-34 and ResNet-50 models
trained on Imagenet-1K and achieved 28.3\% and 25.6\%
Top-1 error rates, respectively. This scheme is our baseline for 2-bits weight and
full-precision activations.
For 2-bits weight and 8-bits activation,
we find work by \citet{PCLpaper1} to achieve the best accuracies reported in literature.
For ResNet-50, \citet{PCLpaper1} obtain 29.24\% Top-1 error. We consider this work to be
our baseline for 2-bits weight and 8-bits activation models.

For 4-bits precision, we find WRPN scheme~\citep{WRPN} to report the highest accuracy.
We implemented this scheme for 4-bits weight and 8-bits activations.
For ResNet-34 and ResNet-50 models trained on Imagenet-1K, we achieve 29.7\% and 28.4\%
Top-1 error rates, respectively.



\subsection{Scheme-A: Joint training of teacher-student networks}
In the first scheme that we investigate, a full-precision teacher network is jointly
trained with a low-precision student network. Figure~\ref{fig:ArchitectureApprenticeFig}
shows the overall training framework. We use ResNet topology for both the teacher and student
network. When using a certain depth for the student network, we pick the teacher network to have
either the same or larger depth.

In~\citet{Compression-Caruana} and~\citet{Hinton-distill}, only the student network trains while
distilling knowledge from the teacher network.
In our case, we jointly train with the rationale that the teacher network would continuously guide
the student network not only with the final trained logits, but also on
what path the teacher takes
towards generating those final higher accuracy logits.

We implement pre-activation version of ResNet~\citep{resnet-v2} in TensorFlow. The training process
closely follows the recipe mentioned in~\citet{FBresnet} - we use a batch size of 256 and
no hyper-parameters are changed from what is mentioned in the recipe. For the teacher network,
we experiment with ResNet-34, ResNet-50 and ResNet-101 as options. For the student network, we
experiment with low-precision variants of ResNet-18, ResNet-34 and ResNet-50.

For low-precision numerics, when using ternary precision we use
the ternary weight network scheme~\citep{TWN} where the weight tensors
are quantized into $\{-1,0,1\}$ with a per-layer scaling coefficient computed based on the mean
of the positive terms in the weight tensor.
We use the WRPN scheme~\citep{WRPN} to quantize weights and activations to 4-bits or 8-bits.
We do not lower the precision of the first layer and the final layer in the apprentice network.
This is based
on the observation in almost all prior works that lowering the precision of
these layers degrades the accuracy dramatically.
While training and during fine-tuning, the gradients are still maintained at full-precision.

\begin{table}[!htb]
  \floatsetup{floatrowsep=qquad, captionskip=2pt}
  \begin{floatrow}
    \ttabbox%
    {\begin{tabularx}{0.85\textwidth}{ccccc}
      \toprule
               & ResNet-18          & ResNet-18             & ResNet-18             & ResNet-18              \\
               & Baseline           & with ResNet-34        & with ResNet-50        & with ResNet-101        \\
               \cmidrule(lr){2-2}\cmidrule(lr){3-3}\cmidrule(lr){4-4}\cmidrule(lr){5-5}
      32A, 32W & 30.4               & 30.2                  & 30.1                  & 30.1                   \\
      32A, 2W  & 33.4               & 31.7                  & 31.5                  & 31.8                   \\
      8A, 4W   & 33.6               & 29.6                  & 29.6                  & 29.9                   \\
    	8A, 2W   & 33.9               & 32.0                  & 32.2                  & 32.4                   \\
      \bottomrule
      \end{tabularx}}
    {\caption[ResNet-18]{Top-1 validation set error rate (\%) on ImageNet-1K for ResNet-18 student
    network as precision of activations (A) and weight (W) changes. The last three columns show error
    rate when the student ResNet-18 is paired
    with ResNet-34, ResNet-50 and ResNet-101.}
    \label{tab:res18}}
  \end{floatrow}
\end{table}%

\begin{wrapfigure}{r}{0.475\textwidth}
\begin{center}
   \includegraphics[width=1\textwidth]{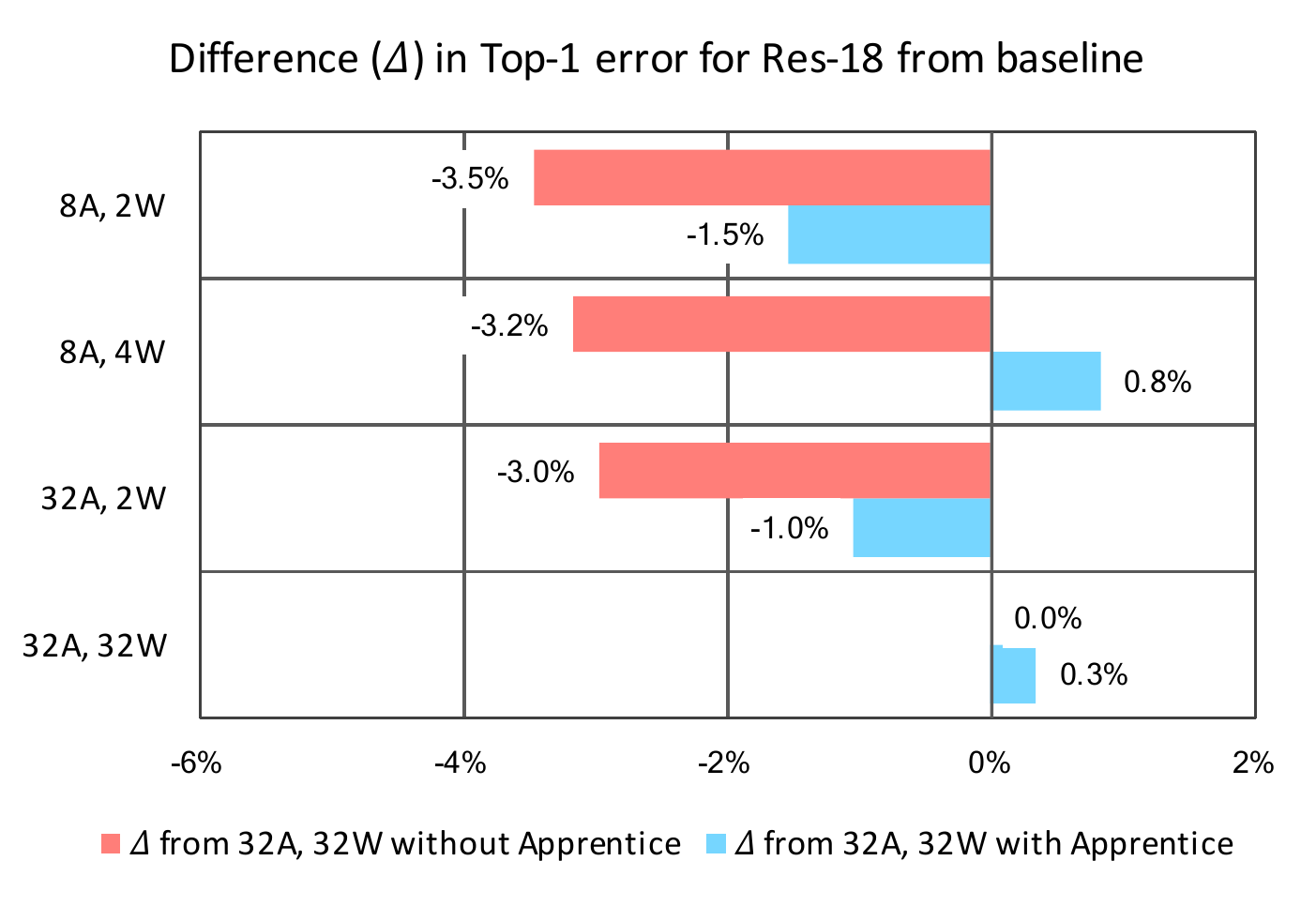}
\end{center}
   \caption{\small Difference in Top-1 error rate for low-precision variants of ResNet-18 with (blue bars) and without (red bars) distillation scheme. The difference is calculated from the accuracy of ResNet-18 with full-precision numerics. Higher \% difference denotes a better network configuration.}
\label{fig:res18}
\end{wrapfigure}

\textbf{Results with ResNet-18}: Table~\ref{tab:res18} shows the effect of lowering precision on the accuracy (Top-1 error) of ResNet-18
with baseline (no teacher) and with ResNet-34, ResNet-50 and ResNet-101 as teachers. In the table, $A$ denotes
the precision of the activation maps (in bits) and $W$ denotes the precision of the weights.
The baseline
Top-1 error for full-precision ResNet-18 is 30.4\%. By lowering the precision without using any help from a teacher network,
the accuracy drops by 3.5\% when using ternary and 4-bits precision (the column corresponding to
``Res-18 Baseline'' in the table).
With distillation based technique, the accuracy of low-precision configurations improves significantly.
In fact, the accuracy of the full-precision ResNet-18 also improves when paired with a larger
full-precision ResNet model (the row corresponding to ``32A, 32W'' in Table~\ref{tab:res18}).
The best full-precision accuracy was achieved with a student ResNet-18 and ResNet-101
as the teacher (improvement
by 0.35\% over the baseline). The gap between full-precision ResNet-18 and the best achieved
ternary weight ResNet-18 is
only 1\% (improvement of 2\% over previous best). With ``8A, 4W'', we find the accuracy of the student ResNet-18 model to
beat the baseline accuracy. We hypothesize regularization with low-precision (and distillation) to be the reason for this.
``8A, 4W'' improving the accuracy beyond baseline figure is only seen for ResNet-18.

Figure~\ref{fig:res18} shows the \textit{difference} in Top-1 error rate achieved by our best
low-precision student networks (when trained under the guidance of a teacher network)
versus not using any help from a teacher network. For this figure, the difference in Top-1 error of the best
low-precision student network is calculated from the baseline full-precision
network (i.e. ResNet-18 with 30.4\% Top-1 error), i.e. we want to see how close a low-precision student network can come
to a full-precision baseline model. We find our low-precision network accuracies to significantly close the gap between
full-precision accuracy (and for some configurations even beat the baseline accuracy).

\citet{Hinton-distill} mention improving the baseline full-precision accuracy when a student network
is paired with a teacher network. They mention improving the accuracy of a small model on MNIST dataset. We
show the efficacy of distillation based techniques on a much bigger model (ResNet) with much larger
dataset (ImageNet).

\begin{table}[!htb]
  \floatsetup{floatrowsep=qquad, captionskip=2pt}
  \begin{floatrow}
    \ttabbox%
    {\begin{tabularx}{0.85\textwidth}{ccccc}
      \toprule
               & ResNet-34          & ResNet-34             & ResNet-34             & ResNet-34              \\
               & Baseline           & with ResNet-34        & with ResNet-50        & with ResNet-101        \\
               \cmidrule(lr){2-2}\cmidrule(lr){3-3}\cmidrule(lr){4-4}\cmidrule(lr){5-5}
      32A, 32W & 26.4               & 26.3                  & 26.1                  & 26.1                   \\
      32A, 2W  & 28.3               & 27.6                  & 27.2                  & 27.2                   \\
      8A, 4W   & 29.7               & 27.0                  & 26.9                  & 26.9                   \\
    	8A, 2W   & 30.8               & 28.8                  & 28.8                  & 28.5                   \\
      \bottomrule
      \end{tabularx}}
    {\caption[ResNet-34]{Top-1 validation set error rate (\%) on ImageNet-1K for ResNet-34 student
    network as precision of activations (A) and weight (W) changes. The last three columns show error
    rate when the student ResNet-34 is paired
    with ResNet-34, ResNet-50 and ResNet-101.}
    \label{tab:res34}}
  \end{floatrow}
\end{table}%

\begin{table}[!htb]
  \floatsetup{floatrowsep=qquad, captionskip=2pt}
  \begin{floatrow}
    \ttabbox%
    {\begin{tabularx}{0.665\textwidth}{cccc}
      \toprule
               & ResNet-50          & ResNet-50             & ResNet-50         \\
               & Baseline           & with ResNet-50        & with ResNet-101   \\
               \cmidrule(lr){2-2}\cmidrule(lr){3-3}\cmidrule(lr){4-4}
      32A, 32W & 23.8               & 23.7                  & 23.5              \\
      32A, 2W  & 26.1               & 25.4                  & 25.3              \\
      8A, 4W   & 28.5               & 25.5                  & 25.3              \\
    	8A, 2W   & 29.2               & 27.3                  & 27.2              \\
      \bottomrule
      \end{tabularx}}
    {\caption[ResNet-50]{Top-1 validation set error rate (\%) on ImageNet-1K for ResNet-50 student
    network as precision of activations (A) and weight (W) changes. The final two columns show error
    rate when the student ResNet-50 is paired
    with ResNet-50 and ResNet-101.}
    \label{tab:res50}}
  \end{floatrow}
\end{table}%

\begin{figure}[!htb]
  \begin{subfigure}[b]{0.495\textwidth}
    \includegraphics[width=1\textwidth]{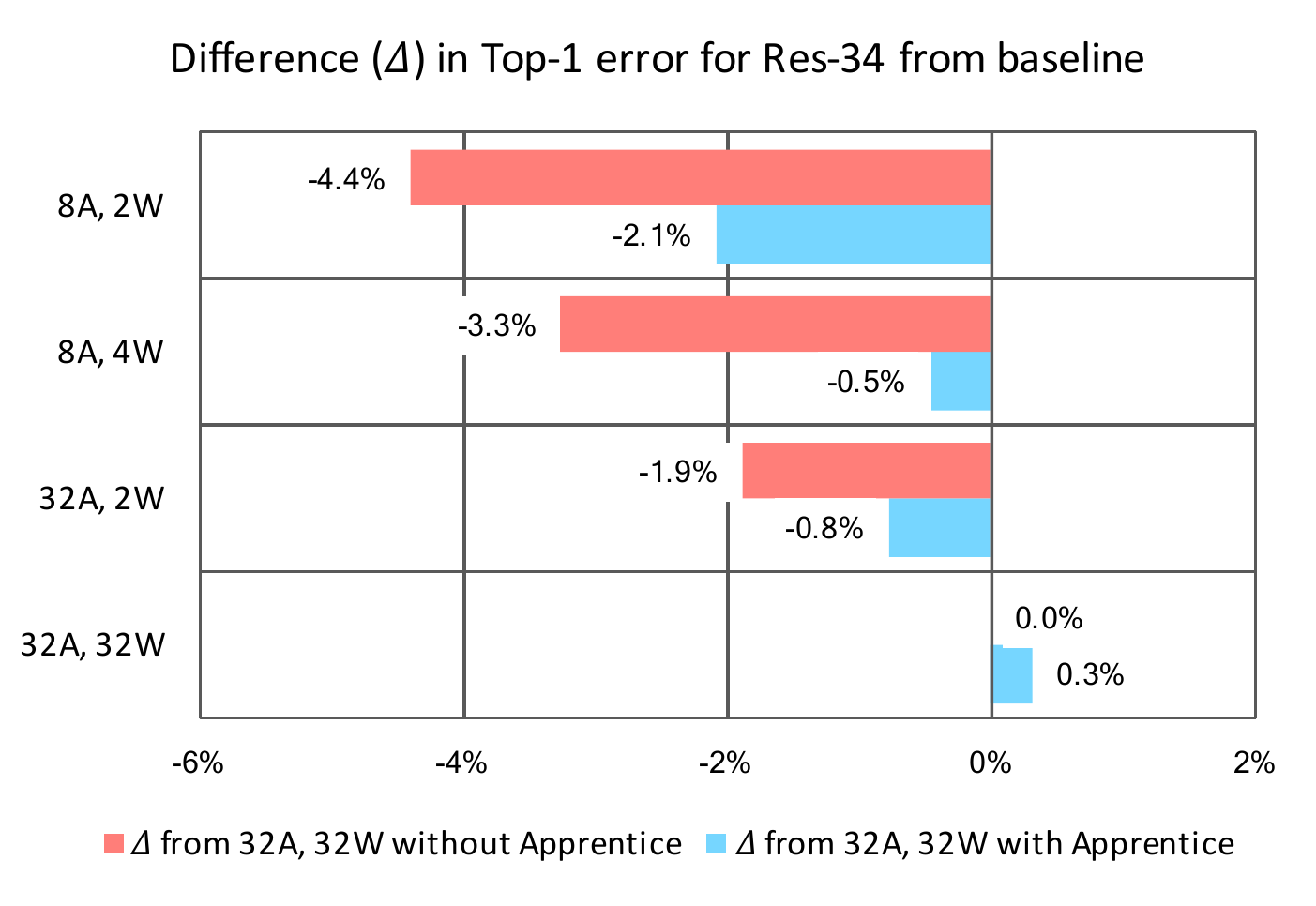}
    \caption{\small Apprentice versus baseline accuracy for ResNet-34.}
    \label{fig:res34}
  \end{subfigure}
  \begin{subfigure}[b]{0.495\textwidth}
    \includegraphics[width=1\textwidth]{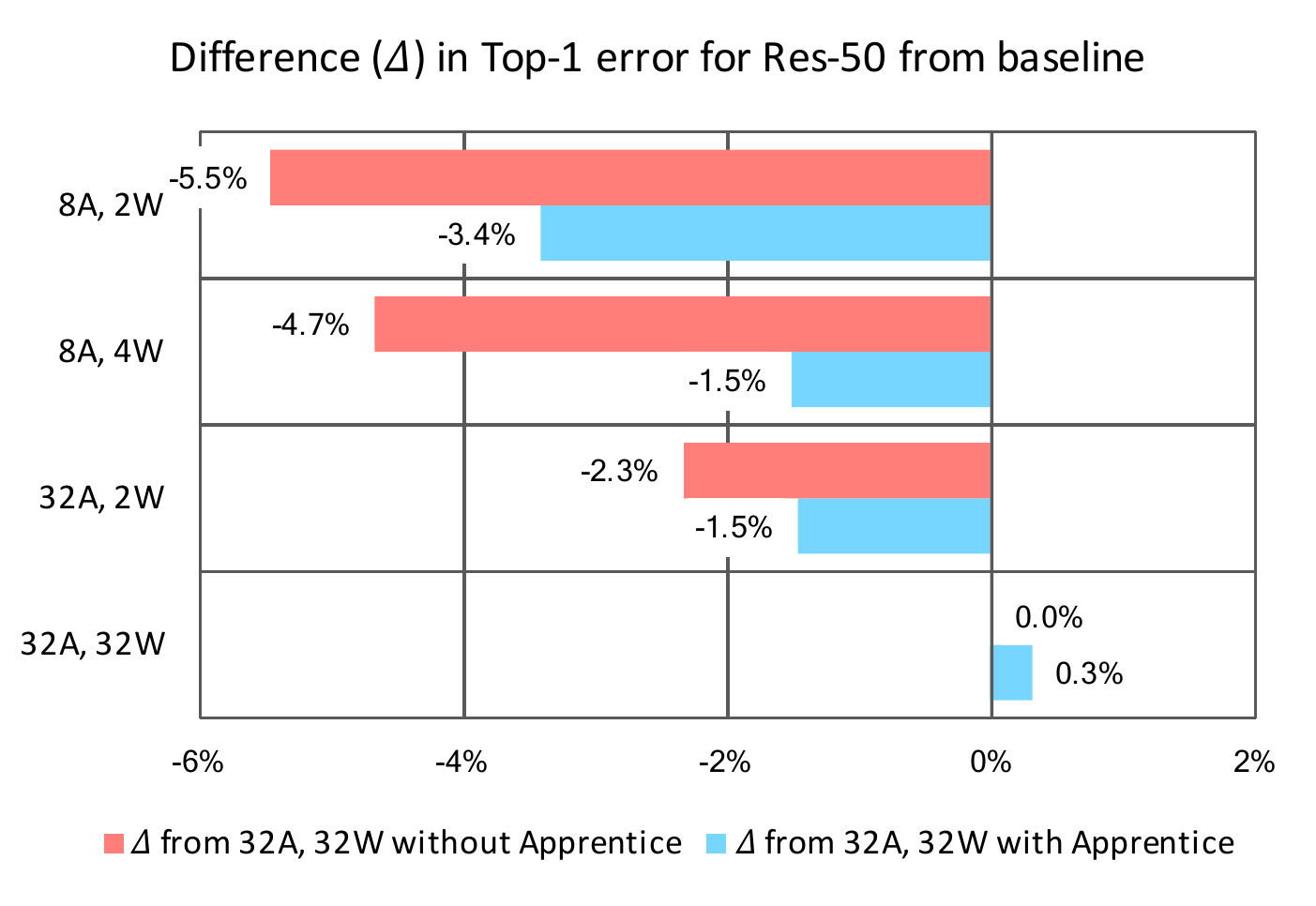}
    \caption{\small Apprentice versus baseline accuracy for ResNet-50.}
    \label{fig:res50}
  \end{subfigure}
  \caption{\small Difference in Top-1 error rate for low-precision variants of ResNet-34 and ResNet-50
  with (blue bars) and without (red bars) distillation scheme.
  The difference is calculated from the accuracy of the baseline network (ResNet-34 for (a) and ResNet-50
  for (b)) operating at full-precision.
  Higher \% difference denotes a better network configuration.}
\end{figure}

\textbf{Results with ResNet-34 and ResNet-50}: Table~\ref{tab:res34} and Table~\ref{tab:res50}
show the effect of lowering precision on the accuracy of ResNet-34 and ResNet-50,
respectively, with distillation based technique.
With a student ResNet-34 network, we use ResNet-34, ResNet-50 and ResNet-101 as teachers.
With a student ResNet-50 network, we use ResNet-50 and ResNet-101 as teachers.
The Top-1 error for full-precision ResNet-34 is 26.4\%. Our best 4-bits weight and 8-bits activation ResNet-34 is within
0.5\% of this number (26.9\% with ResNet-34 student and ResNet-50 teacher). This significantly improves upon the
previously reported error rate of 29.7\%. 4-bits weight and 8-bits activation for ResNet-50 gives us a model that is within
1.5\% of full-precision model accuracy (25.3\% vs. 23.8\%).
Figure~\ref{fig:res34} and Figure~\ref{fig:res50} show
the difference in Top-1 error achieved by our best
low-precision ResNet-34 and ResNet-50 student networks, respectively,
and compares with results obtained using
methods proposed in literature. Our \texttt{Apprentice}
scheme significantly closes the gap between full-precision baseline
networks and low-precision variants of the same networks. In most
cases we see our scheme to better the previously reported
accuracy numbers by 1.5\%-3\%.

\noindent{\textbf{Discussion}}: In scheme-A, we use a teacher network that is always as
large or larger in number of parameters than the student network.
We experimented with a ternary ResNet-34 student network which
was paired with a full-precision ResNet-18. The ternary model for ResNet-34 is
about 8.5x smaller in size compared to the full-precision ResNet-18 model.
The final trained accuracy of the ResNet-34 ternary model with this setup is 2.7\% worse than that
obtained by pairing the ternary ResNet-34 network with a ResNet-50 teacher network. This suggests
that the distillation scheme works only when the teacher network is higher in accuracy
than the student network (and not necessarily bigger in capacity).
Further, the benefit from using a larger teacher network saturates
at some point. This can be seen by picking up a precision point, say ``32A, 2W''
and looking at the error rates along the row in Table~\ref{tab:res18}, \ref{tab:res34} and \ref{tab:res50}.

One concern, we had in the early stages of our investigation, with joint training of a low-precision
small network and a high precision large network was the influence of
the small network's accuracy on the accuracy of the large network. When using the
joint cost function, the smaller network's probability scores are matched with the predictions
from the teacher network. The joint cost is added as a term to the total loss function (equation~\ref{eq1}).
This led us to posit that the larger network's learning capability will be affected by the
inherent impairment in the smaller low-precision network. Further, since the smaller student
network learns form the larger teacher network, a vicious cycle might form where the student network's accuracy
will further drop because the teacher network's learning capability is being impeded.
However, in practice, we did not see this phenomenon occurring - in each case where
the teacher network was jointly
trained with a student network, the accuracy of the teacher network
was always within 0.1\% to 0.2\% of the accuracy of the
teacher network without it jointly supervising a student network. This could be because of our choice of
$\alpha$, $\beta$ and $\gamma$ values.

In Section~\ref{sec:kd}, we mentioned about temperature, $\tau$, for Softmax function
and hyper-parameters $\alpha=1$, $\beta=0.5$ and $\gamma=0.5$.
Since, we train directly on the logits of the teacher network,
we did not have to experiment with the appropriate
value of $\tau$. $\tau$ is required when training on the soft targets produced by the teacher network.
Although we did not do extensive studies experimenting with training on soft targets as opposed to logits,
we did find that $\tau=1$ gives us best results when training on soft targets.
\citet{Hinton-distill} mention that when the
student network is significantly smaller than the teacher network, small values of $\tau$
are more effective than large values. For few of the low-precision configurations,
we experimented with $\alpha=\beta=\gamma=1$, and, $\alpha=0.9$, $\beta=1$ and
$\gamma=0.1$ or $0.3$. Each of these configurations, yielded a lower performance model
compared to our original choice for these parameters.

For the third term in equation~\ref{eq1}, we experimented with a mean-squared error loss
function and also a loss function with logits from both the student and the teacher network
(i.e. $\mathcal{H}(z^T,z^A)$). We did not find any improvement in accuracy compared to
our original choice of the cost function formulation.
A thorough investigation of the behavior of the networks
with other values of
hyper-parameters and different loss functions is an agenda for our future work.


Overall, we find the distillation process to be quite effective in getting us high accuracy low-precision models.
\textit{All our low-precision models surpass previously reported low-precision accuracy figures}.
For example, TTQ scheme achieves 33.4\% Top-1 error
rate for ResNet-18 with 2-bits weight. Our best ResNet-18 model, using scheme-A, with 2-bits weight achieves $\sim$31.5\% error rate,
improving the model accuracy by $\sim$2\% over TTQ. Similarly, the scheme in~\citet{PCLpaper1} achieves 29.2\% Top-1 error
with 2-bits weight and 8-bits activation. The best performing \texttt{Apprentice}
network at this precision achieves 27.2\% Top-1 error.
For Scheme-B and Scheme-C, which we describe next, Scheme-A serves as the new baseline.

\subsection{Scheme-B: Distilling knowledge from a teacher}
In this scheme, we start with a trained teacher network.
Referring back to Figure~\ref{fig:ArchitectureApprenticeFig}, the input image is passed to both the teacher and
the student network, except that the learning with back-propagation happens only in the low
precision student network which is trained from scratch. This is the scheme used by
\citet{Compression-Caruana} and~\citet{Hinton-distill} for training their student networks. In this scheme,
the first term in equation~\ref{eq1} zeroes out and only the last two terms in the equation
contribute toward the loss function.


\begin{figure}[!htb]
\begin{center}
   \includegraphics[width=0.95\textwidth]{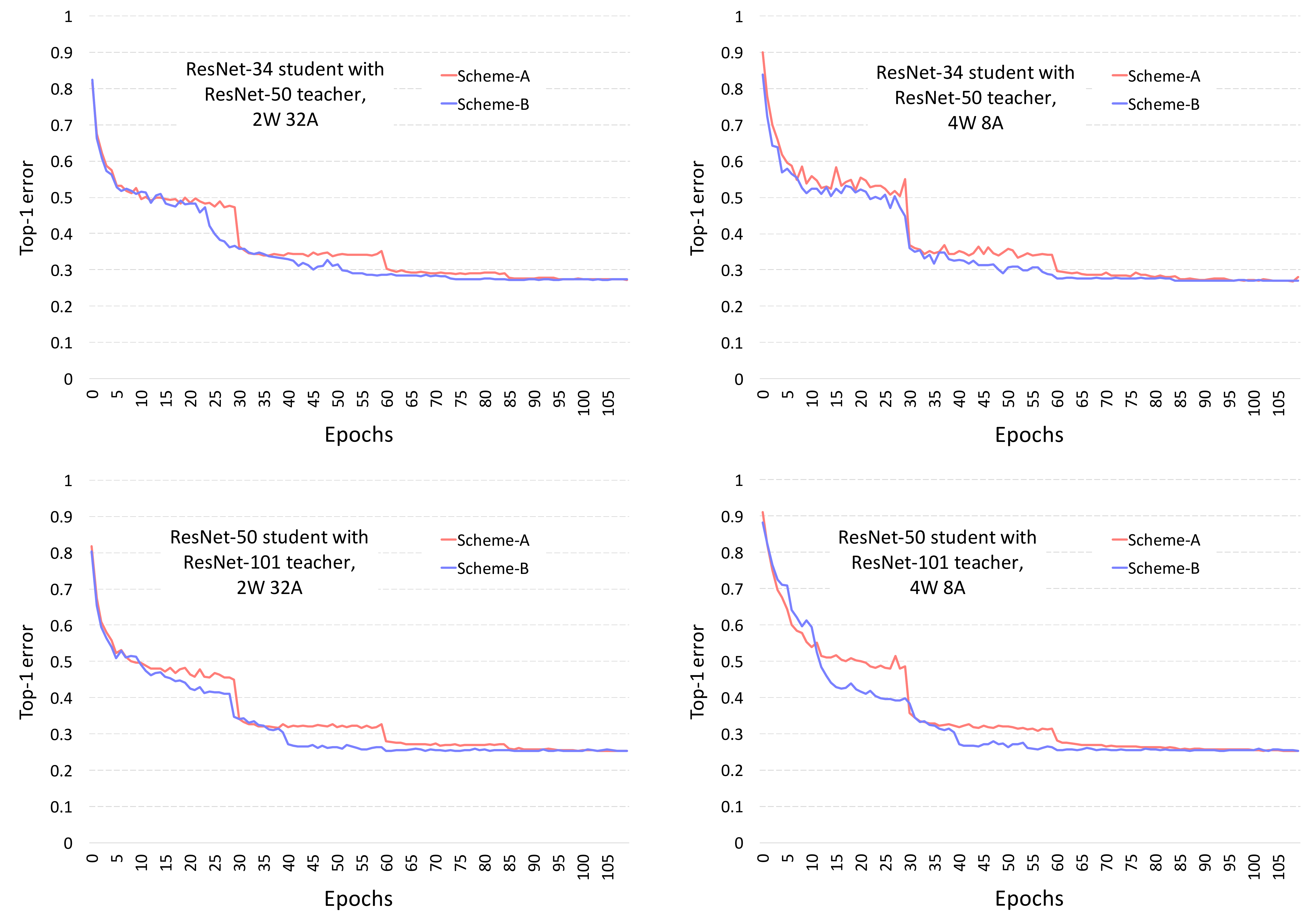}
\end{center}
   \caption{\small Top-1 error rate versus epochs of four student networks using scheme-A and scheme-B.}
\label{fig:curves}
\end{figure}

With scheme-B, one can pre-compute and store the logit values
for the input images on disk and access them during training the
student network. This saves the forward pass
computations in the teacher network. Scheme-B might also help the scenario where a student network
attempts to learn the ``dark knowledge'' from a teacher network that has
already been trained on some private or sensitive data (in addition to the data the student
network is interested in training on).

With scheme-A, we had the hypothesis that the student network would be influenced by not only
the ``dark knowledge'' in the teacher network but also the path the teacher adopts
to learn the knowledge. With scheme-B we find, that the student network gets to similar accuracy numbers as
the teacher network albeit at fewer number of epochs.

With this scheme, the training accuracies are similar to that reported in
Table~\ref{tab:res18}, \ref{tab:res34} and \ref{tab:res50}. The low-precision student
networks, however, learn in fewer number of epochs. Figure~\ref{fig:curves} plots the
Top-1 error rates for few of the configurations from our experiment suite.
In each of these plots, the student network in scheme-B converges around 80th-85th
epoch compared to about 105 epochs in scheme-A. In general, we find
the student networks with scheme-B to learn in about 10\%-20\% fewer
epochs than the student networks trained using scheme-A.

\subsection{Scheme-C: Fine-tuning the student model}
Scheme-C is very similar to scheme-B, except that the student network is primed with full
precision training weights before the start of the training process. At the beginning of the
training process, the weights and activations are lowered and the student network is sort of
fine-tuned on the dataset. Similar to scheme-B, only the final two terms in equation~\ref{eq1}
comprise the loss function and the low-precision student network is trained with back-propagation
algorithm. Since, the network starts from a good initial point, comparatively low learning rate is used
throughout the training process. There is no clear recipe for learning rates (and change of learning rate
with epochs) which works
across all the configurations. In general, we found training with a learning rate of 1e-3 for 10 to 15 epochs,
followed by 1e-4 for another 5 to 10 epochs, followed by 1e-5 for another 5
epochs to give us the best accuracy.
Some configurations run for about 40 to 50 epochs before stabilizing.
For these configurations, we found training using
scheme-B with warm startup (train the student network at full-precision for about
25-30 epochs before lowering the precision) to be equally good.

We found the final accuracy of the models obtained using this scheme to be (marginally) better than
those obtained using scheme-A or scheme-B. Table~\ref{tab:schemeC} shows error rates of few configurations
of low-precision student network obtained using scheme-A (or scheme-B) and scheme-C. For ResNet-50 student network,
the accuracy with ternary weights is further improved by 0.6\% compared to that obtained using scheme-A. Note that
the performance of ternary networks obtained using scheme-A are already state-of-the-art.
Hence, for ResNet-50 ternary networks, 24.7\% Top-1 error rate is the new state-of-the-art. With this,
ternary ResNet-50 is within 0.9\% of baseline accuracy (23.8\% vs. 24.7\%).
Similarly, with 4-bits weight and 8-bits activations, ResNet-50 model obtained using scheme-C is 0.4\% better
than that obtained with scheme-A (closing the gap to be within 1.3\% of full-precision ResNet-50 accuracy).

\begin{table}[!htb]
  \floatsetup{floatrowsep=qquad, captionskip=2pt}
  \begin{floatrow}
    \ttabbox%
    {\begin{tabularx}{0.95\textwidth}{ccc}
      \toprule
               & 32A, 2W                & 32A, 2W            \\
               & with scheme-A or B     & with scheme-C      \\
               \cmidrule(lr){2-2}\cmidrule(lr){3-3}
      ResNet-34 student with ResNet-50 teacher & 27.2 & 26.9 \\
      ResNet-50 student with ResNet101 teacher & 25.3 & 24.7 \\
      \hline\hline
              & 8A, 4W                 & 8A, 4W            \\
              & with scheme-A or B     & with scheme-C      \\
              \cmidrule(lr){2-2}\cmidrule(lr){3-3}
      ResNet-34 student with ResNet-50 teacher & 26.9 & 26.8 \\
      ResNet-50 student with ResNet101 teacher & 25.5 & 25.1 \\
      \bottomrule
      \end{tabularx}}
    {\caption[Scheme-C]{Top-1 ImageNet-1K validation set error rate (\%) with scheme-A and scheme-C for
    ResNet-34 and ResNet-50 student networks with ternary and 4-bits precision.}
    \label{tab:schemeC}}
  \end{floatrow}
\end{table}%

Scheme-C is useful when one already has a trained network which can be fine-tuned using
knowledge distillation schemes to produce a low-precision variant of the trained network.

\subsection{Discussion - ternary precision versus sparsity}
As mentioned earlier, low-precision is a form of model compression. There are many works which target
network sparsification and pruning techniques to compress a model. With ternary precision models, the model
size reduces by a factor of $2/32$ compared to full-precision models.
With \texttt{Apprentice}, we show how one can get a
performant model with ternary precision. Many works targeting network pruning and sparsification target a
full-precision model to implement their scheme. To be comparable in model size to ternary networks,
a full-precision model needs to be sparsified by 93.75\%. Further, to be effective, a sparse model
needs to store a key for every non-zero value denoting the position of the value in the weight tensor.
This adds storage overhead and a sparse model needs to be about 95\% sparse to be at-par in
memory size as a 2-bit model.
Note that ternary precision also has inherent sparsity (zero is a term in the ternary symbol dictionary)
-- we find our ternary models to be about 50\% sparse. In work by~\citet{structured-sparsity} and
\citet{HanPTD}, sparsification of full-precision networks is proposed but the sparsity achieved is less
than 93.75\%. Further, the network accuracy using techniques in
both these works lead to larger degradation in accuracy compared to our ternary models.
Overall, we believe, our ternary precision models to be state-of-the-art not only in
accuracy (we better the accuracy compared to prior ternary precision models)
but also when one considers the size of the model at the
accuracy level achieved by low-precision or sparse networks.

\section{Conclusions}

We present three schemes based on knowledge distillation concept to improve the accuracy
of low-precision networks. Each of the three schemes improve the accuracy of the low-precision
network configuration compared to prior proposals. We motivate the need for a smaller model size
in low batch, real-time and resource constrained inference deployment systems.
We envision the low-precision models produced by our schemes to simplify the inference
deployment process on resource constrained systems and on cloud-based
deployment systems where low latency is a critical requirement.

\newpage
\bibliographystyle{iclr2018_conference}
\bibliography{apprentice_refs}

\newpage
\section{Appendix: Analysis with ResNet on CIFAR-10 dataset}

\begin{figure}[!htb]
  \begin{subfigure}[b]{0.423\textwidth}
    \includegraphics[width=1\textwidth]{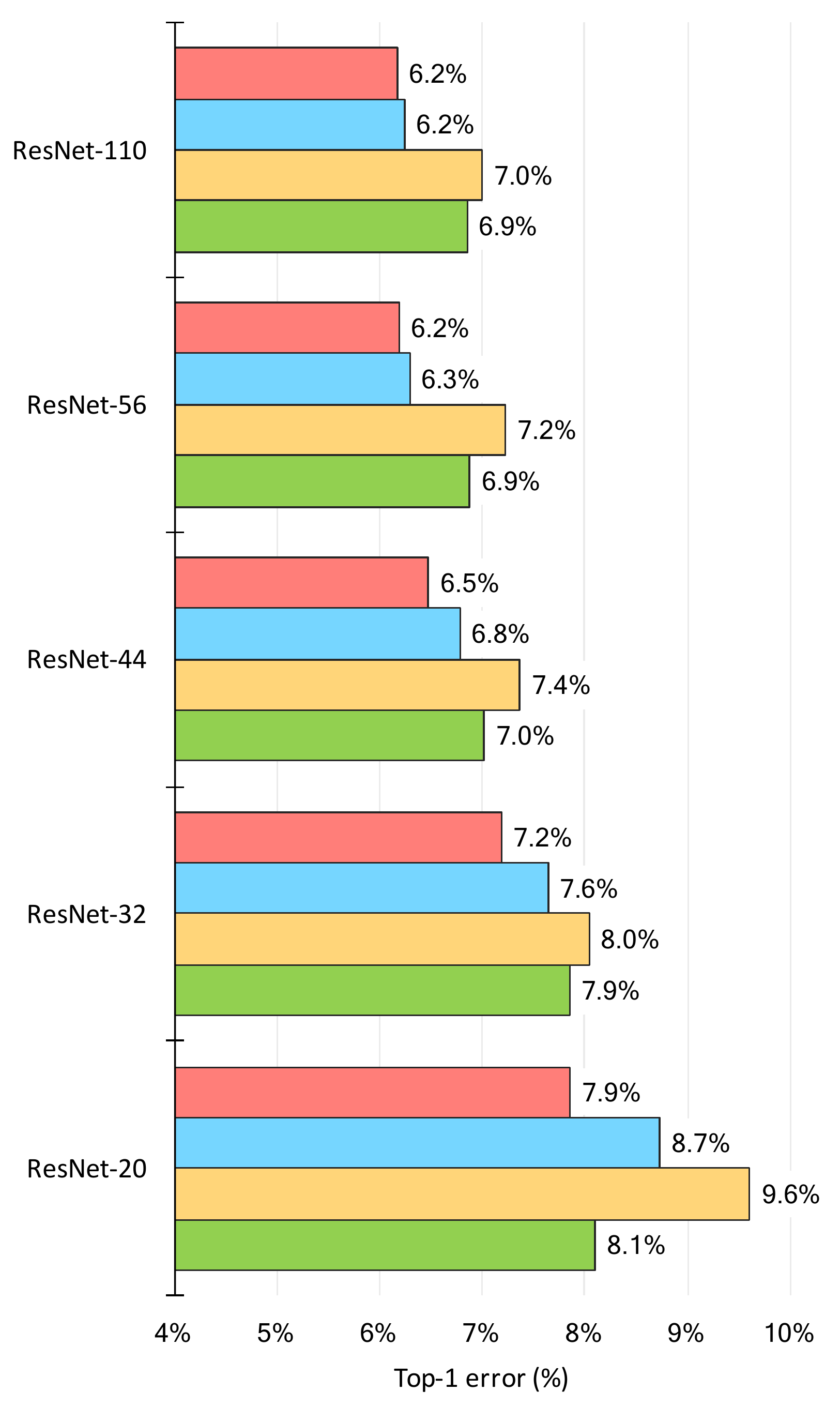}
    \caption{\small Top-1 error without Apprentice scheme.}
    \label{fig:cifar10_withoutApprentice}
  \end{subfigure}
  \begin{subfigure}[b]{0.567\textwidth}
    \includegraphics[width=1\textwidth]{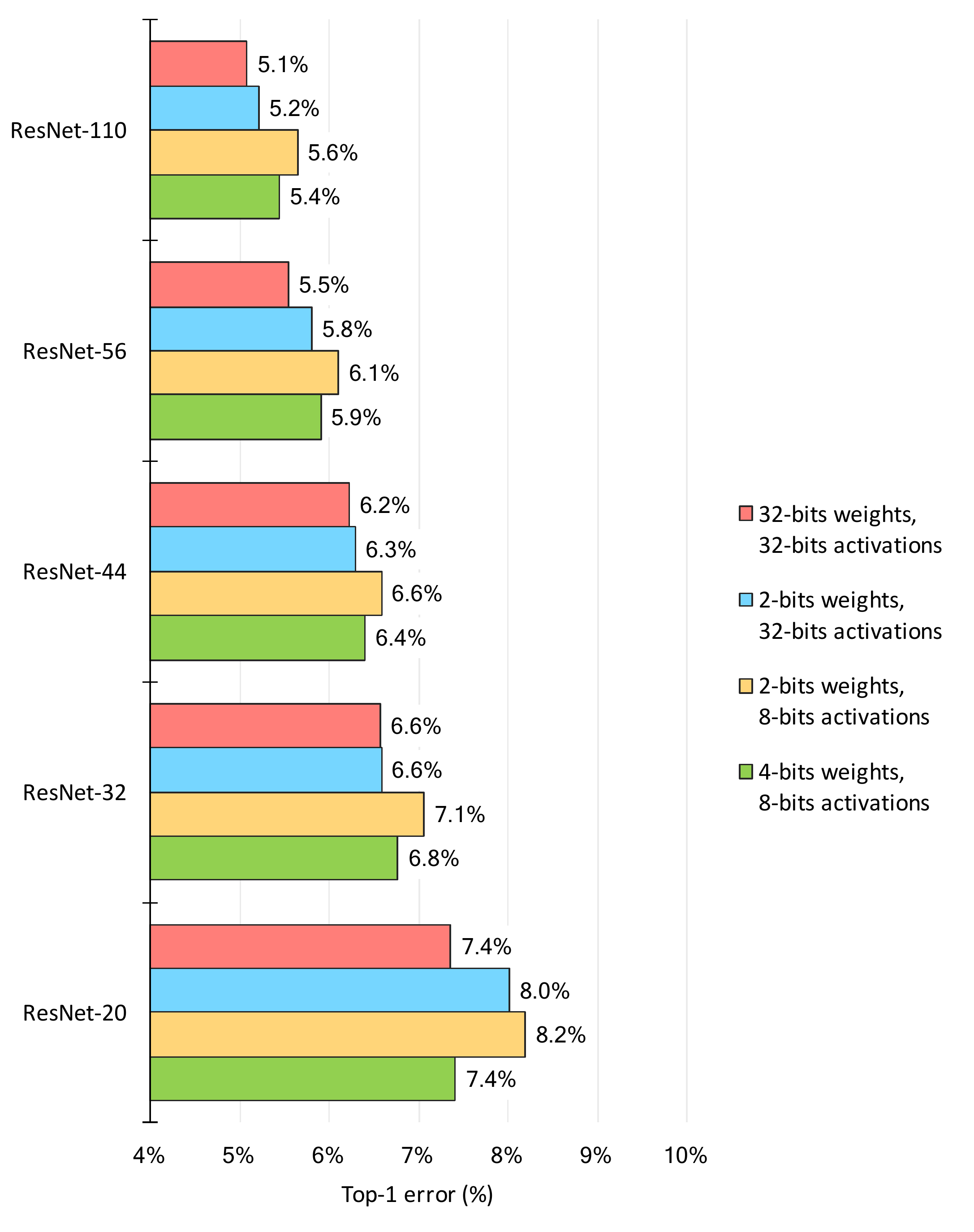}
    \caption{\small Top-1 error \textit{using} Apprentice scheme-A.}
    \label{fig:cifar10_withApprentice}
  \end{subfigure}
  \caption{\small Comparison of various configurations of ResNet on CIFAR-10 with and without Apprentice scheme.}
  \label{fig:cifar10}
\end{figure}

In addition to ImageNet dataset, we also experiment with
\texttt{Apprentice} scheme on CIFAR-10 dataset.
CIFAR-10 dataset~\citep{cifar10} consists of 50K training images and 10K testing
images in 10 classes.
We use various depths of ResNet topology for this study.
Our implemention of ResNet for CIFAR-10 closely follows the configuration
in~\citet{ResNet}.
The network inputs are 32$\times$32 images. The first layer is a 3$\times$3
convolutional layer
followed by a stack of 6$n$ layers with 3$\times$3 convolutions
on feature map sizes 32, 16 and 8;
with 2$n$ layers for each feature map size. The numbers of
filters are 16, 32 and 64 in each set of 2$n$ layers. This is followed by
a global average pooling,
a 10-way fully connected layer and a softmax layer. Thus, in total
there are 6$n$+2 weight layers.

Figure~\ref{fig:cifar10_withoutApprentice} shows the
impact of lowering precision as the depth of ResNet varies. As the network
becomes larger in size, the impact of lowering precision is diminished.
For example,
with ResNet-110, full-precision Top-1 error rate is 6.19\%. At the same depth,
ternarizing the model also gives
similar accuracy (6.24\%). Comparing this with ResNet-20, the gap between
full-precision
and ternary model (2-bits weight and 32-bits activations)
is 0.8\% (7.9\% vs. 8.7\% Top-1 error). Overall, we find that ternarizing a model
closely follows
accuracy of baseline full-precision model. However, lowering both weights and activations
almost always leads to large accuracy degradation. Accuracy of
2-bits weight and 8-bits activation network is 0.8\%-1.6\% worse than full-precision model.
Using \texttt{Apprentice} scheme this gap is considerably lowered.

Figure~\ref{fig:cifar10_withApprentice} shows the impact of lowering precision
when a low-precision (student) network is paired with a full-precision
(teacher) network.
For this analysis we use scheme-A where we jointly train
both the teacher and student network.
The mix of ResNet depths we used for this study
are ResNet-20, 32, 44, 56, 110 and 182.
ResNet-20 student network was paired with deeper ResNets from this
mix, i.e. ResNet-32, 44, 56, 110 and 182 (as five separate experiments).
Similarly, ResNet-44 student network was paired with deeper
ResNet-56 and 110 (as two different set of experiments).
ResNet-110 student network used ResNet-182 as its teacher network.
For a particular ResNet depth, the figure plots the minimum error rate 
across each of the experiments.

We find \texttt{Apprentice} scheme to improve the baseline full-precision
accuracy. The scheme also helps close the gap between the new improved
baseline accuracy and the accuracy when lowering
the precision of the weights and activations.
The gap between 2-bits weight and 8-bits activation network is
now 0.4\%-0.8\% worse than full-precision model.

\end{document}